\documentclass[a4paper]{article}

\usepackage{INTERSPEECH2021}
\usepackage{pgf}
\usepackage{pdfpages}
\usepackage{booktabs}
\usepackage{multirow}
\title{CoDERT: Distilling Encoder Representations with Co-learning for Transducer-based Speech Recognition }

\name{Rupak Vignesh Swaminathan, Brian King, Grant P. Strimel, Jasha Droppo, Athanasios Mouchtaris}
\address{
 Alexa Machine Learning, Amazon.com, USA}
\email{\{swarupak,bbking,gsstrime,drojasha,mouchta\}@amazon.com}

\begin{document}

\maketitle
\begin{abstract}
  We propose a simple yet effective method to compress an RNN-Transducer (RNN-T) through the well-known knowledge distillation paradigm. We show that the transducer's encoder outputs naturally have a high entropy and contain rich information about acoustically similar word-piece confusions. This rich information is suppressed when combined with the lower entropy decoder outputs to produce the joint network logits. Consequently, we introduce an auxiliary loss to distill the encoder logits from a teacher transducer's encoder, and explore training strategies where this encoder distillation works effectively. We find that tandem training of teacher and student encoders with an inplace encoder distillation outperforms the use of a pre-trained and static teacher transducer. We also report an interesting phenomenon we refer to as implicit distillation, that occurs when the teacher and student encoders share the same decoder. Our experiments show 5.37-8.4\% relative word error rate reductions (WERR) on in-house test sets, and 5.05-6.18\% relative WERRs on LibriSpeech test sets.
  
\end{abstract}


\section{Introduction}

As neural network architectures evolve and become more complex, there is a growing need to reduce their footprint, not only for faster runtime and reducing server-side costs but also for on-device deployment.  Following the general trend of on-device machine learning models, streaming automatic speech recognition (ASR) systems on edge devices have become popular in recent years \cite{shangguan2019optimizing,yeh2019transformer}. This has contributed to a paradigm shift in ASR systems, moving away from the traditional factored acoustic and language models to fully end-to-end (E2E) neural-based recognizers, thereby saving significantly on memory and latency \cite{li2019improving, Prabhavalkar2017}. Recent studies have shown that such E2E models can even surpass the performance of conventional server-sized, factored models while being significantly smaller in size \cite{sainath2020streaming,li2020developing}.  

In order to facilitate streaming speech recognition on-device, which enables better latency, privacy and reliability, compression and quantization techniques \cite{xue2013restructuring,prabhavalkar2016compression,NguyenAle2020} have been developed to reduce model sizes and to accelerate inference. Low-rank matrix factorization has been applied successfully \cite{xue2013restructuring, prabhavalkar2016compression} on acoustic models compressing them by a factor of at least 3 times. Pruning methods \cite{wu2020dynamic, frankle2018lottery,ZhenNguyen2021} exploiting sparsity in model weights have also been shown to be promising for ASR footprint reduction. Light recurrent replacements for Long Short-term Memory (LSTM) \cite{hochreiter1997long} cells such as Gated Recurrent Units \cite{ravanelli2018light}, Coupled Input-Forget Gate \cite{greff2016lstm} LSTMs, and Simple Recurrent Units \cite{lei2017simple} have shown negligible degradation in the performance for E2E ASR as well \cite{shangguan2019optimizing}. 


 The Knowledge Distillation (KD) paradigm \cite{hinton2015distilling}, commonly referred to as teacher-student training, has become one of the sought-after techniques for model compression thanks to its architecture-agnostic nature and the ability to take advantage of unsupervised data \cite{parthasarathi2019lessons}. It has been proven to be complementary to other compression approaches \cite{mishra2017apprentice}, yielding additional gains. In this work, we focus on distilling the RNN-T \cite{graves2012sequence} encoder into a smaller, compact encoder. The RNN-T encoder is the most cumbersome part of network, and usually is several folds larger than the decoder. Furthermore, the encoder network processes audio frames that are much longer compared to the label sequence processed by the decoder. Motivated by this, we present CoDERT, a methodology for distilling RNN-T encoder representations, which has not been previously explored. Our contributions in this work are as follows: (1) we show empirically that the RNN-T encoder's high-level representation has rich information about acoustically similar word-pieces, (2) we introduce distillation as an auxiliary task for the RNN-T encoder, distilling from a larger teacher RNN-T encoder, and (3) we demonstrate through experiments that for the encoder distillation to work effectively, the student and the teacher RNN-Ts need to be \textit{co-learned} (trained in tandem) while sharing a common decoder. 



\section{Related Work}
\label{sec:related}

The importance of KD in ASR model compression has been well established for both hybrid acoustic models and E2E models \cite{chebotar2016distilling,kurata2019guiding,liu2021exploiting}. In \cite{kurata2020knowledge}, the authors performed KD from a bi-directional teacher RNN-T into a uni-directional student RNN-T by computing KL-divergence between the outputs of the joint networks of the teacher and student RNN-Ts. This technique can be considered as lattice distillation since the distillation takes place at the end of the joint network's output lattice. The approach does not scale to large batch sizes and output vocabulary sizes, since the distillation loss computation and memory usage become inhibitively expensive \cite{panchapagesan2020efficient}. The authors of \cite{panchapagesan2020efficient} propose a work-around approximation by conglomerating the probabilities of output tokens apart from the blank symbol ($\phi$) and ground-truth symbol ($y_{u+1}$) into a ``garbage bin" and then applying KL-divergence to compute the distillation loss. In \cite{liu2020improving}, the authors propose an auxiliary task for the RNN-T encoder to predict context-dependent graphemic units used in the acoustic model of a traditional hybrid ASR system. This can be viewed as acoustic distillation, which we explore in this work, as opposed to the lattice distillation discussed earlier \cite{kurata2020knowledge, panchapagesan2020efficient}.

 Tandem training of teacher and student models are gaining traction recently \cite{ke2019dual, yu2019universally,yu2021dual}. In \cite{yu2021dual}, a novel RNN-T architecture, which has an encoder working in dual-mode (full-context and streaming) with unique weight sharing strategies, uses \cite{panchapagesan2020efficient} as an \textit{inplace} distillation \cite{yu2019universally} method to distill knowledge from the full-context mode onto the streaming mode. Motivated by \cite{yu2021dual}, we jointly train the student and teacher encoders while sharing a common decoder, which we refer to as \textit{co-learning} in this work. However, compared to \cite{yu2021dual}, our approach focuses only on streaming encoders for both teacher and student models, and more importantly, our distillation loss remains distinct.

\section{Technical Approach}
\label{sec:proposed}
In this section, we review the RNN-T architecture and provide an explanation of our rationale motivating our focus on the RNN-T encoder. We then present the formulation and elucidate our encoder distillation approach in detail, and finally discuss the benefits of the proposed approach.

\subsection{RNN-Transducer}
\label{sec:RNNT}
The RNN-T architecture was introduced in \cite{graves2012sequence} as an extension to the connectionist temporal classification (CTC) framework \cite{Graves06connectionisttemporal} to overcome the conditional independence assumption of CTC. RNN-T consists of an encoder $\mathcal{F}$, which takes audio frame $x_{t}$ as input and produces a high-level representation $h^{enc}_{t}$; a prediction network $\mathcal{G}$ (also referred to as the decoder), which outputs a high-level representation $h^{dec}_{u}$ given the previous non-blank symbol $y^{prev}_u$ as input; and a joint network $\mathcal{J}$ which combines the encoder and decoder representations to produce the token logits. The posterior probability $P^k_{t,u}$ for the output token $k$ at the lattice location $(t,u)$, is computed by applying softmax on the joint network output $h^k_{t,u}$.




 The ground truth label sequence is used to train the decoder (teacher forcing) in the training phase, and the previous non-blank output of the transducer is used as the input to decoder during inference. The joint network strategy in our work is a simple addition followed by a tanh non-linearity. The RNN-T optimization is then formulated as maximizing the sum of all possible alignments $P(y^*|x)$, or equivalently minimizing the negative log-likelihood of $P(y^*|x)$, which is efficiently computed using a forward-backward algorithm as shown in \cite{graves2012sequence}. 


\subsection{Motivation: Why focus on the encoder?}
\label{subsec:motivation}

First, the RNN-T encoder is the largest and most complex sub-network of the model. Unlike the decoder which performs a simple next step prediction task, the encoder's objective is to model speech signals that are highly dynamic with acoustic variability coming from multiple speakers, noises and reverberant conditions. In order to model such variability, the encoder transformation $\mathcal{F}(x)$ requires an extremely high modeling capacity. However, due to the constraints involved in on-device deployment, one typically is required to trade-off accuracy for memory and latency. During inference, the decoder outputs can be cached and the search tree nodes which are expanded can be grouped together for batched computation on the decoder. Meanwhile the encoder, which is orders of magnitude larger must process the audio sequence frame-by-frame which is typically much longer than the label sequence for decoder. The encoder's latency can be improved by techniques such as time reduction \cite{he2019streaming} or bifocal networks \cite{macoskeybifocal} without affecting accuracy, but naively reducing the number of layers and/or hidden units to downsize the encoder can severely impair its modeling capacity \cite{shangguan2019optimizing}. 

Next, we show a few examples of the rich representations learnt by the RNN-T encoder. Since our joint network strategy is an add operation followed by tanh non-linearity, the encoder and decoder output projections are the same dimensions as the number of output tokens. This property allows us to visualize the higher-level representations at the output of encoder in a meaningful fashion; computing $softmax(h^{enc}_{t})$ will produce a probability distribution which can be directly attributed to the output tokens, and an $argmax$ of the result can be considered the output token at time $t$. This is presented in Table \ref{tab:acoustic_similarity}, showcasing the ground truth tokens (word-pieces) and their related acoustically similar word-pieces found in the higher-level representations of the encoder of a fully-trained RNN-T model. 

\begin{table}[hptb!]
\caption{Randomly picked ground truth word pieces corresponding to the words "Alexis", "flurry", and "can you" and their  acoustically similar labels present in the encoder output representation. Underscore indicates beginning of a word.}
\label{tab:acoustic_similarity}
\centering
\begin{tabular}{@{}cc@{}}
\toprule
\begin{tabular}[c]{@{}c@{}}Ground-truth\\ word-pieces\end{tabular} & \begin{tabular}[c]{@{}c@{}}Acoustically similar\\ encoder representation\end{tabular} \\ \midrule
\_alexis                                                           & \_alexa                                                                               \\\midrule
\_flu, r, r, y                                                     & \_fairy                                                                               \\\midrule
\_can, \_you                                                       & \_commute                                                                             \\ \bottomrule
\end{tabular}
\end{table}

\begin{figure}[hbpt!]
  \includegraphics[scale=0.53]{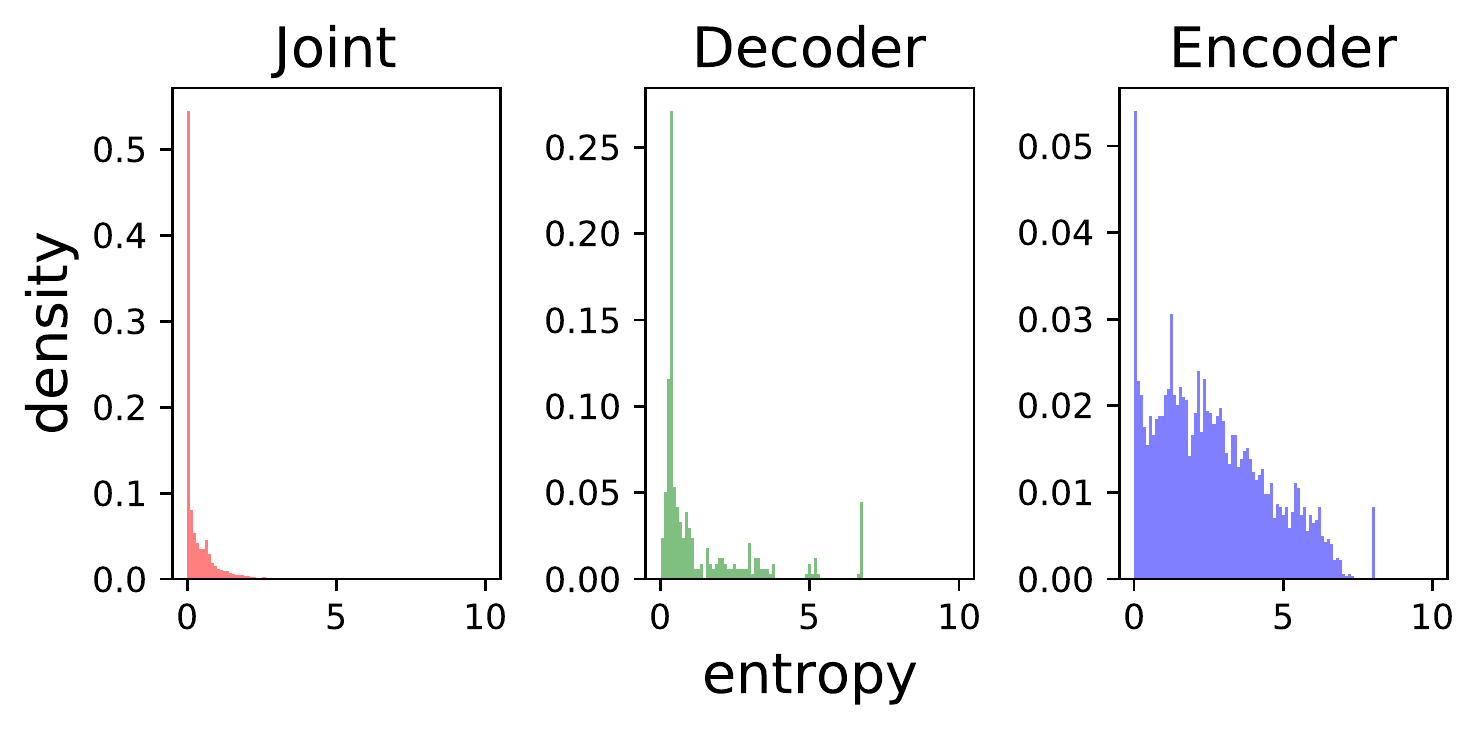}
  \caption{Entropy density of softmax outputs corresponding to the joint (red), decoder (green), and encoder (blue) components of a fully-trained RNN-T with 4K word-pieces.}
  \label{fig:EntHist} 
\end{figure}

Last, we visualize the entropy of the output of each of the components in the RNN-T, and demonstrate that the encoder naturally has a much higher entropy while the decoder's output are highly predictable. Figure \ref{fig:EntHist} shows the density of entropy at the output of joint, decoder, and encoder components, computed over a randomly chosen training batch. Based on the joint network entropy, one concludes that when encoder outputs are combined with that of the decoder, much of the valuable information about the relative probabilities between the output tokens of the encoder is lost. The relative distance of how close or far two sounds are to each other, and a high output entropy, are important characteristics for an effective distillation \cite{hinton2015distilling}; the RNN-T encoder output naturally contains such information as shown in Table \ref{tab:acoustic_similarity} and Figure \ref{fig:EntHist}, respectively.


\vspace{-0.4cm}
\subsection{Formulation: Distilling the high-level representations}
Our encoder distillation formulation is simple yet powerful and easily reproducible. During training, we introduce an auxiliary task for the student encoder to match the logits of the teacher encoder's logits. We experimented with different loss functions ($L_2$, softmax cross-entropy and cosine-loss) and found $L_2$ loss to yield better results.

\begin{equation}
\label{eq:distill}
    \mathcal{L}_{distill} = \| h^{enc}_{1:t}(S) - h^{enc}_{1:t}(T) \|^2
\end{equation}

\noindent where $(S)$ and $(T)$ denote components belonging to student and teacher, respectively.

\begin{figure}[hbpt!]
  \includegraphics[scale=0.65]{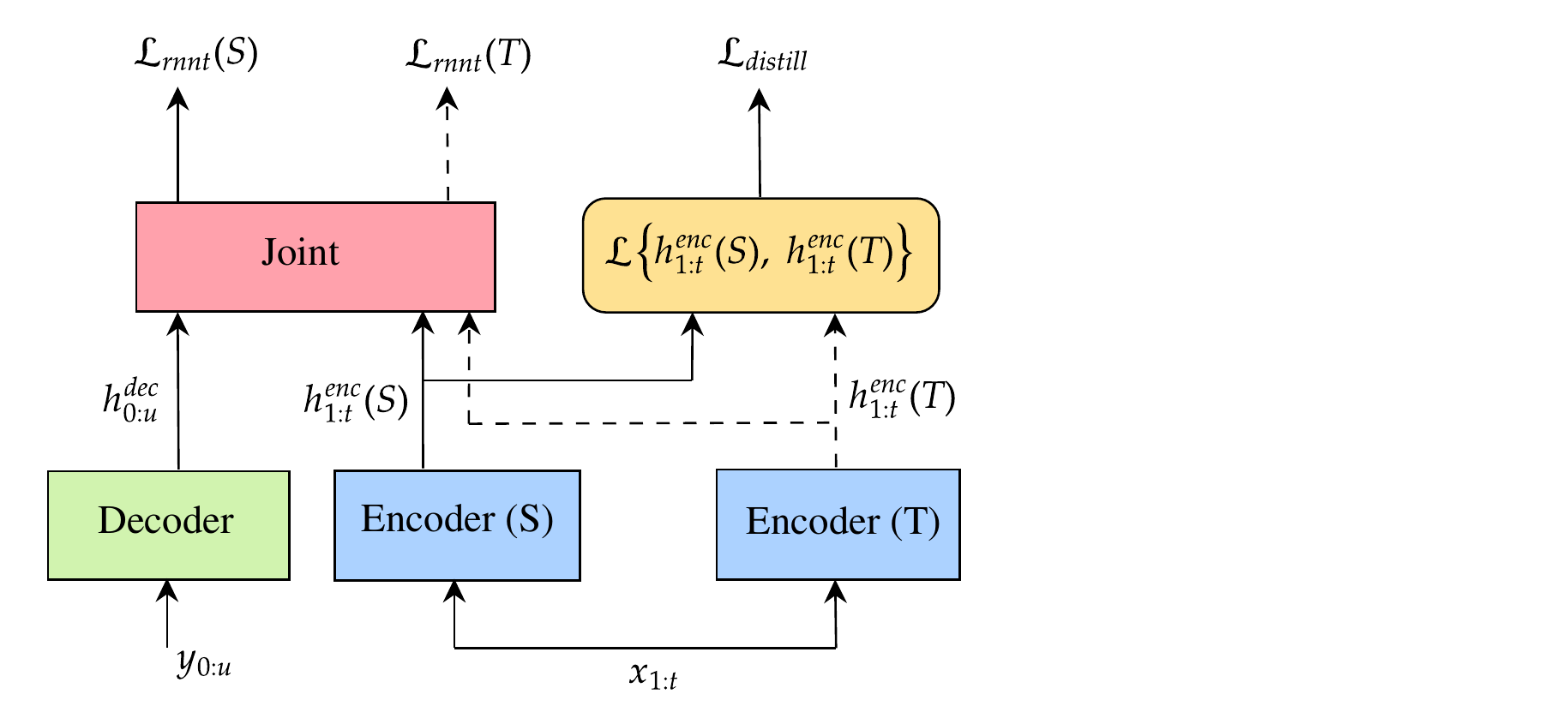}
  \caption{CoDERT: Proposed method for distilling encoder representations. The student (S) and teacher (T) encoders share a common decoder, and are updated through the sum of the RNN-T losses (co-learned). Additionally, the student encoder is updated through the distillation loss presented in Equation \ref{eq:distill}.}
  \label{fig:training} 
\end{figure}

As discussed in Section \ref{sec:results} in detail, we find that in order for the distillation to work effectively, the student and teacher encoders need to share the same decoder (see Figure \ref{fig:training}). Unlike traditional distillation, where the teacher is pre-trained and its parameters are frozen, we update the teacher parameters and perform the distillation \textit{inplace} \cite{yu2019universally,yu2021dual}. The final loss is expressed as,

\begin{equation}
\label{eq:total_loss}
    \mathcal{L} = \mathcal{L}_{rnnt}(S) + \mathcal{L}_{rnnt}(T) + \lambda \mathcal{L}_{distill}
\end{equation}

\noindent where $\lambda$ is the distillation weight, chosen through hyper-parameter optimization on a development set.

\vspace{-0.4cm}
\subsection{Benefits: An information-preserving distillation}

 The output of the joint network is a 4-dimensional tensor of shape ${B \times T \times U \times V}$, where $B$, $T$, $U$, and $V$ correspond to batch size, maximum length of audio frames, maximum length of labels, and the vocabulary size, respectively. The traditional KL-divergence loss between teacher and student RNN-T joint outputs becomes infeasible for production quality training choices of $B$, $T$, $U$, and $V$, since the loss computation and memory bandwidth to maintain such large multi-dimensional tensors become highly impractical on today's machines. An attempted work-around proposed in \cite{panchapagesan2020efficient} is to consider only three probabilities $P^{y+1}_{t,u}$, $P^{\phi}_{t,u}$ and $1-P^{y+1}_{t,u}-P^{\phi}_{t,u}$ (shortened as $P^{rest}_{t,u}$), which are then used in the KL-divergence based distillation loss computation. However, in bucketing the remaining probabilities to compute $P^{rest}_{t,u}$, information about the relative probabilities at the output of joint network is lost. Furthermore, the relative probability of encoder output tokens is not preserved at the joint output as shown in Section \ref{subsec:motivation}. In contrast, the proposed method leverages our findings that most of the information is in the encoder, and distills over the entire ${B \times T \times V}$ space, thereby circumventing the scaling barrier without having to compromise on the relative probabilities. Finally, our method does not preclude one from additionally applying the lattice distillation proposed in \cite{panchapagesan2020efficient}, so it can be considered a complementary approach. 

\vspace{-0.2cm}
\section{Experiments}
\label{sec:expts}
In this section, we discuss briefly about the experimental setup for training and evaluating our encoder distillation approach.

\vspace{-0.2cm}
\subsection{Model architecture}
Our teacher encoder consists of 5 LSTM layers with 1024 hidden units each, followed by a fully-connected output projection layer of 4001 dimensions. The student encoder consists of 4 LSTM layers with 640 units each making it 35\% of the teacher encoder's size. The encoders use Log-Mel filterbank energies (LFBE)  of 64-dimensions, with a window size of 25ms and hop size of 10ms and downsampled by 3 to have a frame rate of 30ms. The encoders also have a time reduction component \cite{he2019streaming} after the second layer which reduces the number of audio frames by a factor of 2. The decoder size is kept constant for both student and teacher models to ensure comparable analysis in our experiments. The decoder network consists of an input embedding of 512-dimensions, a 2-layer LSTM with 1024 hidden units and an output vocab projection of 4001-dimensions. The joint network consists of an addition operation on the encoder and decoder outputs followed by a tanh non-linearity. The output tokens modeled by the transducer are 4001 word pieces which include the blank symbol $\phi$. 

\vspace{-0.2cm}
\subsection{In-house data experimental setup}
\label{sec:alexa_data}
We use a 10K-hour in-house collection of de-identified Alexa English data to investigate our distillation method. We use 60 hours of audio as the development set for hyper-parameter optimization and selecting final model checkpoints, and a 60-hour test set to evaluate our models. Additionally, we also use a long-form ASR utterances test set (\textit{long}) and a tail distribution low-frequency utterances test set (\textit{tail}) in our evaluation. Our learning rate schedule policy has a warm up stage where the learning rate linearly increases from 1e-7 to 5e-4 (\textit{warm up}) for 3K steps, stays constant at 5e-4 for 35K steps (\textit{hold}), and then exponentially decays to final learning rate of 1e-5 at 75K steps. In these experiments, the convergence is achieved usually in between 85K - 100K steps. We report the relative word error rate reduction (WERR) over the baseline model in Table \ref{tab:CaseI} trained without distillation, throughout all experiments with in-house data. We use a beam size of 6 during inference. 

\vspace{-0.2cm}
\subsection{LibriSpeech experimental setup}
 For LibriSpeech \cite{panayotov2015librispeech} experiments, we remove the time reduction component in the encoders and introduce a dropout rate of 30\% for the decoder. The vocabulary size is 4096 (+1 blank symbol), hence the fully-connected output projection from the encoder/decoder becomes 4097.  The learning rate is also modified accordingly; the \textit{warm up} and \textit{hold} stages are shortened to 1K steps and 20K steps, respectively. The models usually converge around 80-95K steps in these experiments. A beam size of 16 is used during decoding. 
 
 \vspace{-0.2cm}
 \section{Results and Discussion}
\label{sec:results}
First, using the experimental setup explained in Section \ref{sec:alexa_data}, we demonstrate that the encoder distillation is ineffective when the student and teacher models are not co-learned, and do not share a common decoder. In other words, the teacher and student are two separate RNN-Ts, and the encoder representation of a pre-trained, static teacher RNN-T is used in distillation. The teacher RNN-T is trained on the same data and its parameters are frozen during the student RNN-T training. From the results in Table \ref{tab:CaseI}, we notice only minimal improvements in the range of 0.9-1.61\% relative WERR. 

Next, we show that in order for the encoder distillation to work effectively, the decoder must be shared across the teacher and student encoders as seen in Figure \ref{fig:training}. We use Equation \ref{eq:total_loss} to compute the total loss, and co-learn both the teacher and student parameters. From the results in Table \ref{tab:CaseI}, we can notice that consistent improvements across all test sets are obtained even before the distillation loss is introduced (i.e., $\lambda$=0.0). Upon introducing the distillation loss (i.e., $\lambda$=1.0), there is additional improvements  in the range of 1.58-3.47\%. The long-form ASR test set (\textit{long}) benefits the most (8.4\% improvement), suggesting the distilled encoder is able to retain long-range acoustic contexts. It is worth noting that the co-learned teacher (Teacher\textsubscript{Co}) performs at the same level of the standalone-trained teacher except on the tail distribution test set.

\vspace{-0.15cm}
\begin{table}[hbpt!]
\caption{On in-house test sets: relative WERR (\%) reported over the Baseline model with smaller encoder trained without distillation. Vanilla student model (row 2) distilled from a pre-trained, static teacher shows minimal improvements, whereas the CoDERT trained student ($\lambda$=1.0) shows significant improvement on all test sets.}
\label{tab:CaseI}
\centering
\begin{tabular}{@{}lllll@{}}
\toprule
Model & \begin{tabular}[c]{@{}l@{}}Encoder \\ Params\end{tabular} & test(\%) & long(\%) & tail(\%) \\ \midrule
\begin{tabular}[c]{@{}l@{}}Baseline\\ \end{tabular} & 16M & - & - & - \\ \midrule
\begin{tabular}[c]{@{}l@{}}Student\\ \end{tabular} & 16M & 1.61 & 1.73 & 0.9 \\ \midrule
\begin{tabular}[c]{@{}l@{}} Teacher\end{tabular} & 46M & 20.16 & 21.15 & 16.45 \\ \midrule
\begin{tabular}[c]{@{}l@{}} CoDERT\\\end{tabular} &  &  &  &  \\ 
\begin{tabular}[c]{@{}l@{}} \hspace{0.2cm} $\lambda$=$0.0$\end{tabular} & 16M & 4.03 & 4.93 & 3.79 \\ 
\textbf{\begin{tabular}[c]{@{}l@{}} \hspace{0.2cm} \boldmath$\lambda$=$1.0$ \end{tabular}} & \textbf{16M} & \textbf{6.45} & \textbf{8.4} & \textbf{5.37} \\ 
\begin{tabular}[c]{@{}l@{}} Teacher\textsubscript{Co}\end{tabular} & 46M & 20.16 & 21.15 & 15.82 \\
\bottomrule
\end{tabular}
\end{table}




Next, we study the effect of distilling top-k logits of highest magnitudes. Preserving only top-k logits for distillation is a common practice in hybrid acoustic model distillation \cite{movsner2019improving}, to remove unwanted noise in the logits. However, as shown in Table \ref{tab:topk}, it is noticeable that our encoder distillation works best, especially on \textit{long} and \textit{tail} test sets, when all 4001 logits are used in distillation. Our findings corroborate with the earlier discussions on KD taking advantage of high entropy at the encoder output and advantage of distilling over the entire ${B \times T \times V}$ space instead of preserving only the top-k logits. 



\vspace{-0.15cm}
\begin{table}[hbpt!]
\caption{Results corresponding to the CoDERT student ($\lambda$=1.0): relative WERR numbers for distilling using only the top-k logits of the encoder.}
\label{tab:topk}
\centering
\begin{tabular}{@{}cccc@{}}
\toprule
top-k value & test (\%) & long (\%) & tail (\%)\\ \midrule
5 & 4.04 & 4.93 & 3.16 \\ \midrule
50 & 4.83 & 5.79 & 4.43 \\ \midrule
500 & 6.45 & 7.53 & 4.43 \\ \midrule
\textbf{4001 (all)} & \textbf{6.45} & \textbf{8.4} & \textbf{5.37} \\ \bottomrule
\end{tabular}
\end{table}

\begin{figure}[hbpt!]
  \includegraphics[scale=0.4]{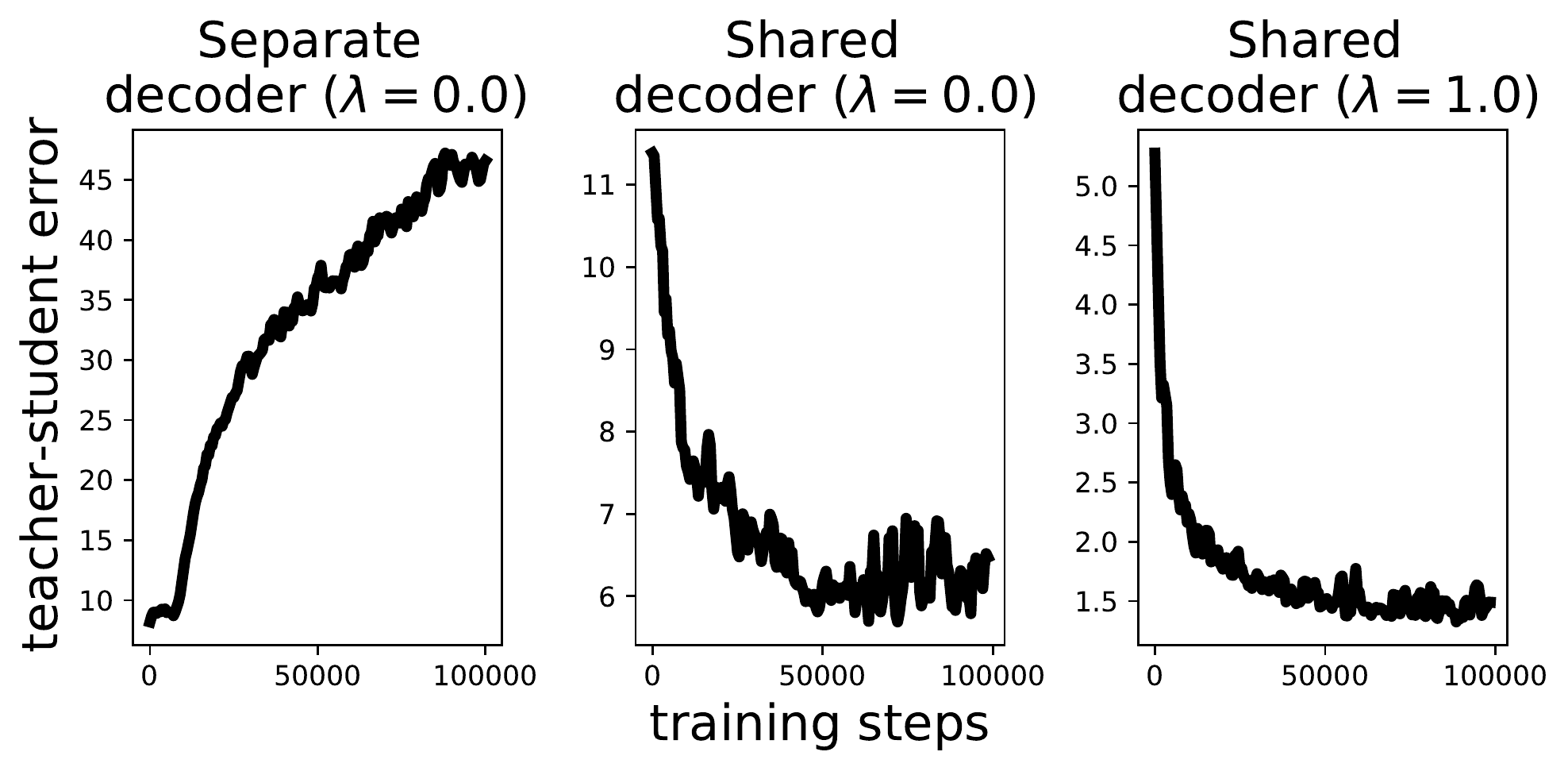}
  \caption{Error between teacher and student encoder logits in (a) separate decoder without distillation (left), (b) shared decoder without distillation (center), and (c) shared decoder with distillation. }
  \label{fig:ImpDist} 
\end{figure}

\vspace{-0.1cm}
\noindent\textbf{Implicit Distillation: }It is evident from results reported in Table \ref{tab:CaseI} that using shared decoder and co-learning increases the distillation efficiency. This is because the decoder learns its own unique representations and greatly influences the final joint network outputs (see Figure \ref{fig:EntHist}), and in learning the alignment \cite{ghodsi2020rnn}. Hence, the encoder distillation is nearly futile without sharing the decoder with the teacher. Interestingly, we discover a phenomenon we call \textit{implicit distillation} (Figure \ref{fig:ImpDist}, center plot), where the output between the teacher and student encoders are implicitly minimized without introducing the distillation loss ($\lambda=0.0$) when the decoder is shared. The teacher-student error is further driven down by introducing the distillation loss.

\vspace{-0.1cm}

\vspace{-0.2cm}
Finally, we select the best settings from in-house data experiments (shared decoder, co-learning, and full-logit distillation instead of only top-k logits) moving forward with LibriSpeech data. 
The results presented in Table \ref{tab:libri} once again indicate the competency of CoDERT across all dev and test sets on LibriSpeech. The co-learned model without distillation shows no improvement on test-clean, and a modest improvement on test-other (2.29\%), but upon introducing the distillation loss, one can notice significant WER reductions on both test-clean (6.18\%) and test-other (5.05\%).



\vspace{-0.1cm}
\begin{table}[hbtp!]
\caption{On LibriSpeech: Absolute (abs) WER and relative (rel) WERR numbers on LibriSpeech data for baseline model trained without distillation and models trained through CoDERT.}
\label{tab:libri}
\centering
\begin{tabular}{p{1.1cm}p{0.25cm}p{0.25cm}p{0.25cm}p{0.25cm}p{0.25cm}p{0.25cm}p{0.25cm}p{0.25cm}}
\toprule
\multirow{2}{*}{Model} & \multicolumn{2}{c}{dev-clean} & \multicolumn{2}{c}{dev-other} & \multicolumn{2}{c}{test-clean} & \multicolumn{2}{c}{test-other} \\ \cmidrule(l){2-9} 
 & abs. & rel. & abs. & rel. & abs. & rel. & abs. & rel. \\ \midrule 
Baseline & 9.6 & - & 21.1 & - & 9.7 & - & 21.8 & - \\ \midrule
CoDERT &  &  &  &  &  &  &  &  \\ 
 \hspace{0.2cm}$\lambda$=0.0 & 9.1 & 5.2 & 20.5 & 2.84 & 9.7 & 0.0 & 21.3 & 2.29 \\ 
\hspace{0.1cm} \boldmath$\lambda$=$1.0$ & \textbf{8.9} & \textbf{7.29} & \textbf{19.8} & \textbf{6.16} & \textbf{9.1} & \textbf{6.18} & \textbf{20.7} & \textbf{5.05} \\ 
Teacher\textsubscript{Co} & 7.9 & 17.7 & 18.8 & 10.9 & 8.3 & 14.43 & 19.5 & 10.55 \\ \bottomrule
\end{tabular}
\end{table}



\vspace{-0.5cm}
\section{Conclusion}
\label{sec:conc}
In this work, we explored distilling the high-level representations of the RNN-T encoder and investigated its effectiveness in building a RNN-T model with a compact encoder. We studied training strategies for encoder distillation and learned that the method works at its best when the student RNN-T shares its decoder with a teacher model that is trained in tandem. We reported an interesting implicit distillation behaviour that occurs during the tandem training. Our results showed significant improvements across various test sets of both real-world and open-source data, demonstrating the efficacy of the proposed technique. 
\vspace{-0.2cm}
\section{Acknowledgements}
We thank Ariya Rastrow, Gautam Tiwari, Harish Arsikere and Maurizio Omologo for helpful discussions and feedback.

\bibliographystyle{IEEEtran}
\bibliography{mybib}


\end{document}